\documentclass[11pt]{article}

\usepackage[margin=1in]{geometry}
\usepackage{latexsym}
\usepackage{booktabs}
\usepackage{enumitem}
\usepackage{hyperref}
\usepackage{xcolor}
\usepackage{url}
\usepackage{amsmath}

\hypersetup{
  colorlinks=true,
  linkcolor=blue,
  citecolor=blue,
  urlcolor=blue
}

\title{Naamah: A Large Scale Synthetic Sanskrit NER Corpus via DBpedia Seeding and LLM Generation}

\author{
Annarao Kulkarni \and Akhil Rajeev P\\
Centre for Development of Advanced Computing (C-DAC), Bangalore\\
\texttt{\{akhil.rajeev, annarao\}@cdac.in}
}

\date{}

\begin{document}
\maketitle

\begin{abstract}
The digitization of classical Sanskrit literature is impeded by a scarcity of annotated resources, particularly for Named Entity Recognition (NER). While recent methodologies use generic Large Language Models (LLMs) for data augmentation, these approaches remain prone to error and often lack the reasoning depth required for classical grammar. In this work, we introduce Naamah, a high quality silver standard Sanskrit NER dataset comprising 102,942 sentences. We propose a methodology that combines entity extraction from DBpedia with the generative capabilities of a 24B parameter hybrid reasoning model to create grammatically natural and synthetically diverse training data. We use this dataset to benchmark two transformer architectures: the massive multilingual XLM-RoBERTa and the parameter efficient IndicBERTv2. Our experiments reveal a key insight: while both models scale well with synthetic data, IndicBERTv2 qualitatively outperforms XLM-RoBERTa in entity identification and classification. On a fixed split of 92,647 train and 10,295 validation examples, IndicBERTv2 achieves the best validation F1 of 0.9615, outperforming XLM-RoBERTa's 0.9506 while remaining substantially lighter for deployment. We demonstrate that the generic tokenizer of XLM-RoBERTa fractures Sanskrit terms, whereas the domain adapted tokenizer of IndicBERTv2 preserves semantic integrity.
\end{abstract}


\section{Introduction}
The Naamah dataset is publicly available at \url{https://huggingface.co/datasets/akhil2808/Naamah}.

Sanskrit is central to South Asian intellectual history, yet modern NLP resources for Sanskrit remain sparse relative to contemporary high resource languages. Extracting structured information from this corpus constitutes a significant challenge in digital humanities. Named Entity Recognition (NER) serves as the foundational step for downstream tasks such as knowledge graph construction, relation extraction, digital philology, and historical prosopography. However, developing NER systems for Sanskrit is complicated by two primary factors: intrinsic linguistic complexity and a scarcity of annotated resources.

Sanskrit is a morphologically rich language characterized by extensive agglutination and inflection. Unlike English, where word order largely determines syntax, Sanskrit relies on a complex system of case markers, or \textit{vibhakti}. A single named entity, such as Rama, can manifest in more than 24 surface forms depending on its syntactic role. Furthermore, entities are often merged phonetically with adjacent words via sandhi, obscuring their boundaries. Standard string matching techniques or rigid rule based systems often fail to capture this variance in unseen contexts.

The resource bottleneck is significant. Manual annotation requires high level domain expertise, making the creation of gold standard corpora slow and expensive. Existing datasets are often small, domain specific, or affected by severe class imbalance. Current approaches to overcoming this include cross lingual transfer, where labels are projected from high resource languages such as English. However, projection methods introduce significant alignment noise due to structural mismatches. Similarly, using generic LLMs for generation often yields errors because such models lack domain specific grounding for Indic scripts.

In this paper, we leverage a hybrid reasoning model optimized for Indic languages to bridge this data gap. Our contributions are threefold:
\begin{enumerate}[leftmargin=*]
    \item \textbf{DBpedia mining strategy:} We detail a methodology for extracting diverse entity seeds from DBpedia using structured queries, ensuring broad coverage of persons, locations, and organizations.
    \item \textbf{The Naamah corpus:} We introduce a silver standard dataset of 102,942 Sanskrit sentences generated through this model and refined through heuristic preprocessing. This method bypasses rigid grammar templates, allowing varied syntactic structures to emerge naturally.
    \item \textbf{Benchmarking insights:} We provide a comparative analysis of XLM-RoBERTa Base and IndicBERTv2 on a fixed split of 92,647 training examples and 10,295 validation examples. We demonstrate that for classical languages, domain aligned tokenization is more critical than raw model scale.
\end{enumerate}

\section{Related Work}
\subsection{Challenges in Cross Lingual Projection}
A common approach to low resource NER involves projecting annotations from a source language to the target language via parallel corpora. For instance, the Naamapadam dataset \cite{mhaske2023naamapadam} uses the Samanantar corpus to generate NER data for 11 Indic languages. Linguistic mismatches such as divergence in word order and the lack of direct equivalents for Sanskrit case markers lead to alignment errors. Parallel alignment errors can propagate directly into label quality, especially with inflected forms. Our work circumvents this by generating data directly in the target language structure.

\subsection{Sanskrit Computational Linguistics and NER}
Rule based paradigms have dominated traditional Sanskrit processing. Tools like the Sanskrit Heritage Reader \cite{goyal2016sandhi} excel at morphological analysis and segmentation. However, these tools lack the probabilistic flexibility required to disambiguate named entities in complex contexts where ambiguity is resolved through broader sentence semantics. While deep learning has been applied to segmentation \cite{hellwig2020segmentation}, contextual NER remains under explored. Early efforts in Sanskrit NER largely relied on rule based heuristics and dictionary lookups, which naturally struggle with out of vocabulary terms and extensive sandhi \cite{murthy2008rule}. Subsequent attempts have explored statistical models such as Conditional Random Fields on limited, domain specific corpora \cite{bhargava2016ner}. This research gap is further widened by the lack of inclusion in foundational datasets; for instance, Naamapadam, the primary large scale repository for NER in Indic languages, does not currently include Sanskrit.

\subsection{Sanskrit Digital Resources}
The development of robust NLP models for classical languages heavily depends on the availability of digitized texts and lexical frameworks. Several notable efforts have laid the groundwork for Sanskrit digital humanities. The Digital Corpus of Sanskrit \cite{hellwig2010dcs} provides an extensively annotated corpus for morphological and lexical analysis, while the Goettingen Register of Electronic Texts in Indian Languages serves as a comprehensive repository of machine readable foundational texts. Additionally, lexical resources such as IndoWordNet \cite{bhattacharyya2010indowordnet} offer valuable semantic linkages. While these digital resources are valuable for philological research, grammar formulation, and basic NLP tasks, they generally lack the dense, large scale semantic annotations required for training modern deep learning based NER systems. This scarcity directly underscores the need for the synthetic data generation pipeline proposed in this work.

\subsection{Synthetic Data Generation}
Data augmentation is a standard technique in low resource NLP \cite{ding2020daga}. The current trend relies heavily on generic LLMs \cite{wang2023gptner}, which can generate incorrect grammatical structures in low resource languages. Our work uses an LLM optimized specifically for Indic scripts, offering a domain grounded generative alternative.

\section{Automated Entity Extraction from DBpedia}
A critical challenge in synthetic data generation is ensuring the diversity of the lexicon. If the model observes only a handful of traditional names during training, it will simply memorize those tokens rather than learning the morphological context of a named entity. To address this, we propose leveraging DBpedia, a large scale multilingual knowledge base.

\subsection{Knowledge Base Structure}
DBpedia organizes knowledge as a graph of triples using the Resource Description Framework. This structure allows researchers to programmatically filter entities based on their ontology classes.

\subsection{Extraction Methodology}
By using SPARQL, we extracted a broad spectrum of entities targeting three primary categories: person, location, and organization.

To ensure morphological variety, the extraction included a diverse mix of both classical Indian entities and global entities, including modern international locations and foreign political figures transliterated into Devanagari script. Embedding transliterated names such as Giacomo Libera or Manfred Hake alongside traditional Sanskrit entities prevents downstream NER models from relying on lexical familiarity, forcing them to learn the underlying syntactic patterns and case markers that designate an entity in a Sanskrit sentence.

\section{The Naamah Corpus}
Using the vetted entity lists, we developed our dataset. We shifted from a deterministic logic approach to an LLM driven generative pipeline to maximize syntactic fluidity.

\subsection{Language Model Pipeline}
Instead of relying on rigid, pre-programmed morphological engines that often struggle with the fluid nature of Sanskrit syntax, we used Sarvam-M, a 24-billion-parameter hybrid reasoning model heavily optimized for Indic languages.

\paragraph{Generation process.}
The model was prompted to incorporate specific entity seeds from our DBpedia extraction into semantically coherent Sanskrit sentences. This generative approach allows for the natural emergence of appropriate case endings and phrasing that mimics authentic text better than brittle template only generation, yielding a wider variety of syntactic structures and inflectional realizations.

\paragraph{Preprocessing and heuristics.}
To ensure that the dataset could serve as a reliable silver standard, the raw output underwent a Python based preprocessing layer. Generated candidates are filtered using rule based checks for token-label consistency, malformed output, and ambiguous boundaries. After filtering, we retain 102,942 high quality silver standard examples.

\subsection{Dataset Characteristics and Statistics}
The final dataset consists of 102,942 sentences structured in JSONL format, providing a substantial corpus for training and evaluation. It uses the standard BIO tagging scheme to represent entity boundaries. These tags are mapped to numeric identifiers via a label2id dictionary to facilitate model processing: \texttt{O}: 0, \texttt{B-PER}: 1, \texttt{I-PER}: 2, \texttt{B-ORG}: 3, \texttt{I-ORG}: 4, \texttt{B-LOC}: 5, and \texttt{I-LOC}: 6.

We performed a statistical analysis of the generated corpus. The dataset relies on a highly diverse vocabulary, featuring 123,923 unique tokens across a total volume of 732,267 tokens. The average sentence length is 7.11 tokens.

\begin{table}[t]
\centering
\begin{tabular}{lr}
\toprule
Statistic & Value \\
\midrule
Total Sentences & 102,942 \\
Train Split & 92,647 \\
Validation Split & 10,295 \\
Unique Tokens & 123,923 \\
Total Tokens & 732,267 \\
Average Sentence Length & 7.11 \\
\bottomrule
\end{tabular}
\caption{Core statistics of the Naamah corpus and split configuration used in experiments.}
\label{tab:stats}
\end{table}

\begin{table}[t]
\centering
\begin{tabular}{lr}
\toprule
Entity Class & Count (B tags) \\
\midrule
Person (PER) & 90,452 \\
Location (LOC) & 22,290 \\
Organization (ORG) & 14,655 \\
\midrule
Total Entities & 127,397 \\
\bottomrule
\end{tabular}
\caption{Entity distribution in Naamah.}
\label{tab:entities}
\end{table}

\section{Experimental Setup}
We benchmarked two state of the art transformer models on our dataset to evaluate their capacity to learn from synthetic Sanskrit data.

\subsection{Models}
\begin{enumerate}[leftmargin=*]
    \item \textbf{XLM-RoBERTa Base:} This serves as a strong multilingual baseline. It uses a large vocabulary of 250k tokens and is often the default choice for low resource languages. However, its generic training data includes very little classical Sanskrit.
    \item \textbf{IndicBERTv2 MLM Only:} This provides an Indic focused compact alternative. It uses parameter sharing to reduce size to approximately 130 MB, making it suitable for edge deployment. Its vocabulary is optimized for Indic scripts.
\end{enumerate}

\subsection{Tokenization Strategy and De-Sandhi}
Sanskrit is highly agglutinative, and authentic texts often feature virtually unbounded string sequences due to complex sandhi, or phonetic fusions across word boundaries. While traditional Sanskrit NLP pipelines heavily rely on explicit de-sandhi preprocessing to separate these compound structures before tagging, our methodology evaluates the capacity of modern transformer tokenizers to handle raw, unsplit text natively. Rather than applying a dedicated de-sandhi tool, we rely on the subword tokenization algorithms inherent to XLM-RoBERTa and IndicBERTv2 to implicitly segment these agglutinated forms.

To handle the resulting fragmented subwords during NER training, we employed a label-first alignment strategy. The BIO tag is assigned only to the first subtoken of an entity, and subsequent subtokens are masked with the ignore index $-100$, a standard strategy for token classification with subword tokenizers. This forces the model to predict the entity type based on the root stem while implicitly learning the suffix structure and sandhi fusions.

\subsection{Training Configuration}
Both models are fine tuned with Hugging Face Trainer. XLM-RoBERTa is trained for 3 epochs; IndicBERTv2 is trained for 4 epochs. Batch size is 16. Learning rates are $2 \times 10^{-5}$ for XLM-RoBERTa and $3 \times 10^{-5}$ for IndicBERTv2.

\section{Results and Analysis}
\subsection{Quantitative Results}
Both models achieved strong convergence on the synthetic test set. On a fixed validation split of 10,295 examples, IndicBERTv2 achieves the best validation F1 of 0.961451, outperforming XLM-RoBERTa's 0.950581.

\begin{table}[t]
\centering
\begin{tabular}{lrr}
\toprule
Metric & XLM-RoBERTa & IndicBERTv2 \\
\midrule
Precision & 0.949766 & 0.959563 \\
Recall & 0.951396 & 0.963345 \\
F1 Score & 0.950581 & 0.961451 \\
Accuracy & 0.985695 & 0.988897 \\
Validation Loss & 0.057814 & 0.054086 \\
\bottomrule
\end{tabular}
\caption{Validation performance on Naamah (10,295 examples). IndicBERTv2 achieves the strongest overall NER quality.}
\label{tab:results}
\end{table}

\subsection{Training Dynamics}
XLM-RoBERTa converges in 3 epochs with gradual F1 gains from 0.9366 to 0.9506. IndicBERTv2 converges in 4 epochs with stronger final validation F1, rising from 0.9536 to 0.9615. This indicates improved fit to Sanskrit entity morphology under the same data regime.

\subsection{Qualitative Analysis: Tokenizer Model Fit}
While aggregate scores demonstrate the viability of both models, qualitative error inspection shows a recurring issue for XLM-RoBERTa on inflected forms where suffix fragments receive unstable labels. We tested the models on sentences containing entities and structures not explicitly prevalent in the training data.

\subsubsection{Failure Mode Analysis: Tokenizer Fragmentation}
We observed a recurring failure mode in XLM-RoBERTa with complex agglutinated terms. For the input \textit{Kuruksetre}, meaning ``in Kurukshetra'':
\begin{itemize}[leftmargin=*]
    \item \textbf{IndicBERTv2 output:} \textit{Kuruksetre}, correctly classified as Location.
    \item \textbf{XLM-RoBERTa output:} \textit{Kuruksetra}, classified as Location, plus \textit{e}, incorrectly classified as Organization.
\end{itemize}

XLM-RoBERTa's multilingual tokenizer, which is not optimized for Indic scripts, fractures the word into a root and a suffix. The self attention mechanism treats the suffix as a separate token. Without specific pretraining on Sanskrit morphology, XLM-RoBERTa falsely predicts that this dangling suffix is an Organization. In contrast, IndicBERTv2 handles these subword transitions coherently, recognizing that the suffix modifies the root and maintaining the Location tag across the entire span. IndicBERTv2 is more consistent on complete entity spans, supporting the hypothesis that Indic oriented tokenization better preserves morphological cues crucial for Sanskrit NER.

\section{Discussion}
A significant advantage of the proposed approach is that it effectively bypasses the requirement of manual annotation for named entities. Naamah demonstrates a practical path to scale labeled data for classical languages where expert annotation is scarce. While silver standard data does not replace gold corpora, it provides a strong foundation for pretraining and supervised transfer.

The results suggest that tokenizer language alignment is a primary factor for Sanskrit NER, often more influential than parameter count alone. For practitioners, this implies that compact domain adapted models can outperform larger multilingual encoders when script and morphology differ substantially from pretraining distributions.

\section{Limitations and Future Work}
Naamah is synthetic and inherits biases from source entities, prompting templates, and filtering heuristics. As a preliminary study, this work opens several avenues for critical improvement to transition from a silver standard to a production grade system.

\subsection{Complex Sandhi Resolution}
Future work will include targeted stress testing for complex sandhi. While the generative LLM approach captures basic morphological fusions naturally, authentic Sanskrit literature is dominated by highly complex sandhi across multiple words.

\subsection{Gold Standard Evaluation}
Future work will focus on evaluating manually annotated texts through hybrid training with an expert validated gold subset. We plan to benchmark the pretrained IndicBERTv2 model against excerpts from classical and contemporary Sanskrit texts. This addresses the critical lack of Sanskrit support in modern NER datasets like Naamapadam and the conceptual absence of organization entities in classical corpora by adapting the annotation schema for historical contexts.

\section{Conclusion}
In this work, we introduced Naamah, a large scale, silver standard Sanskrit Named Entity Recognition dataset comprising 102,942 synthetically generated sentences. To overcome the critical scarcity of annotated classical Sanskrit texts, we developed a data generation pipeline that combined structured entity mining from DBpedia with the generative capabilities of a 24-billion-parameter hybrid reasoning LLM optimized for Indic languages. This methodology allowed us to bypass rigid, rule based grammatical templates, resulting in a morphologically diverse and syntactically natural corpus.

We subsequently used this dataset to benchmark two distinct transformer architectures on a fixed split of 92,647 training and 10,295 validation examples. Our evaluations demonstrated that the parameter efficient IndicBERTv2 achieved the highest validation F1 score of 0.9615, outperforming the larger multilingual XLM-RoBERTa score of 0.9506. Crucially, our qualitative analysis revealed that generic multilingual tokenizers frequently fracture complex, agglutinated Sanskrit terms, leading to misclassification. In contrast, domain adapted tokenization successfully preserves entity boundaries and semantic integrity. Ultimately, Naamah provides a robust foundational resource for advancing Sanskrit computational linguistics, demonstrating that language aligned tokenization and targeted synthetic generation can effectively bridge the data gap for low resource classical languages.

\section*{Data Availability}
The dataset introduced in this paper is available at \url{https://huggingface.co/datasets/akhil2808/Naamah}.

\bibliographystyle{plain}

\end{document}